\documentclass{article}

\usepackage[accepted]{icml2011}
\usepackage{fancyhdr}
\usepackage{algorithm}
\usepackage{algorithmic}
\usepackage{amsmath}
\usepackage{amssymb}
\usepackage{graphicx}
\usepackage{times}
\usepackage{paralist}

\newcommand{\namecite}[1]{\citeauthor{#1}~(\citeyear{#1})} %
\newcommand{\ind}{\mathbf{I}}
\newcommand{\E}{\mathbf{E}}
\newcommand{\Var}{\mathbf{Var}}

\newcommand{\A}{\mathcal{A}}
\newcommand{\X}{\mathcal{X}}
\newcommand{\vr}{\vec{r}}
\newcommand{\hrho}{\hat{\rho}}
\newcommand{\hp}{\hat{p}}
\newcommand{\hV}{\hat{V}}
\newcommand{\eps}{\varepsilon}
\renewcommand{\rho}{\varrho}

\newcommand{\scalefigs}{0.9}
\newcommand{\fig}[1]{Fig.~\ref{fig:#1}}

\newcommand{\set}[1]{\{#1\}}
\newcommand{\card}[1]{\lvert#1\rvert}
\newcommand{\abs}[1]{\lvert#1\rvert}
\newcommand{\bigAbs}[1]{\lvert#1\rvert}

\newcommand{\BigAbs}[1]{\Bigl\lvert#1\Bigr\rvert}
\newcommand{\given}{\mathbin{\vert}}
\newcommand{\bigBracks}[1]{\bigl[#1\bigr]}
\newcommand{\BiggBracks}[1]{\Biggl[#1\Biggr]}
\newcommand{\BigBracks}[1]{\Bigl[#1\Bigr]}
\newcommand{\bigParens}[1]{\bigl(#1\bigr)}

\newcommand{\BiggParens}[1]{\Biggl(#1\Biggr)}

\newcommand{\DR}{\text{\mdseries\upshape DR}}
\newcommand{\DM}{\text{\mdseries\upshape DM}}
\newcommand{\IPS}{\text{\mdseries\upshape IPS}}

\newtheorem{theorem}{Theorem}
\newcommand{\eq}[1]{Eq.~\eqref{eq:#1}}
\renewcommand{\sec}[1]{Section~\ref{sec:#1}}
\newcommand{\thm}[1]{Theorem~\ref{thm:#1}}

\DeclareMathOperator*{\argmax}{argmax}

\begin{document}

\twocolumn[
\icmltitle{Doubly Robust Policy Evaluation and Learning}

% It is OKAY to include author information, even for blind
% submissions: the style file will automatically remove it for you
% unless you've provided the [accepted] option to the icml2011
% package.
\icmlauthor{Miroslav Dud\'ik}{mdudik@yahoo-inc.com}
%\icmladdress{}
\icmlauthor{John Langford}{jl@yahoo-inc.com}
\icmladdress{Yahoo! Research, New York, NY, USA 10018}
\icmlauthor{Lihong Li}{lihong@yahoo-inc.com}
\icmladdress{Yahoo! Research, Santa Clara, CA, USA 95054}

% You may provide any keywords that you 
% find helpful for describing your paper; these are used to populate 
% the "keywords" metadata in the PDF but will not be shown in the document
\icmlkeywords{contextual bandit, partial label, multiclass classification}

\vskip 0.1in
]

\begin{abstract}
  We study decision making in environments where the reward is only
  partially observed, but can be modeled as a function of an action
  and an observed context. This setting, known as contextual bandits,
  encompasses a wide variety of applications including health-care
  policy and Internet advertising. A central task is evaluation of a new
  policy given historic data consisting of contexts, actions and
  received rewards. The key challenge is that the past data typically
  does not faithfully represent proportions of actions taken by a new
  policy. Previous approaches rely either on models of rewards
  or models of the past policy.  The former are plagued by a large bias
  whereas the latter have a large variance.
  
  In this work, we leverage the strength and
  overcome the weaknesses of the two approaches by applying
  the \emph{doubly robust} technique to the problems of
  policy evaluation and optimization. We prove that this
  approach yields accurate
  value estimates when we have \emph{either}
  a good (but not necessarily consistent) model of rewards
  \emph{or} a good (but not necessarily consistent) model of past policy. Extensive empirical
  comparison demonstrates that the doubly robust approach uniformly
  improves over existing
  techniques, achieving both lower variance in value estimation and better
  policies.  As such, we
  expect the doubly robust approach to become common practice.  
\end{abstract}

\section{Introduction}
\label{sec:introduction}

We study decision making in environments where we receive feedback
only for chosen actions. For example, in Internet
advertising, we find only whether a user clicked on some of the
presented ads, but receive no information about the ads that were not presented. In health care, we only find out success rates for patients who received
the treatments, but not for the alternatives.  Both
of these problems are instances of \emph{contextual
  bandits}~\cite{EXP4,Epoch-Greedy}. The context refers to
additional information about the user or patient. Here, we focus on the
%non-interactive
offline version: we assume access to historic data, but
no ability to gather new data~\cite{Langford08Exploration,ESII}.

Two kinds of approaches address offline learning in contextual
bandits. The first, which we call the \emph{direct
method} (DM), estimates the reward function from given data and uses
this estimate in place of actual reward to evaluate the policy value
on a set of contexts. The second kind, called \emph{inverse propensity
score} (IPS)~\cite{HorvitzTh52}, uses importance weighting to correct for the incorrect
proportions of actions in the historic data. The first approach requires
an accurate model of rewards, whereas the second approach requires an
accurate model of the past policy. In general, it might be difficult
to accurately model rewards, so the first assumption can be too
restrictive. On the other hand, it is usually possible to model the
past policy quite well. However, the second kind of approach often
suffers from large variance especially when the past policy differs
significantly from the policy being evaluated.

In this paper, we propose to use the technique of \emph{doubly robust} (DR)
estimation to overcome problems with the two existing
approaches.
Doubly robust (or doubly protected)
estimation~\cite{CasselSaWr76,Robins94Estimation,RR95,LD04,KangSc07}
is a statistical approach for
estimation from incomplete data with an important property: if \emph{either one}
of the two estimators (in DM and IPS) is correct, then the estimation is unbiased.
This method thus increases the chances of drawing reliable inference.

For example, when conducting a survey, seemingly ancillary questions
such as age, sex, and family income may be asked.  Since not everyone
contacted responds to the survey, these values along with census
statistics may be used to form an estimator of the probability of a
response conditioned on age, sex, and family income.  Using importance
weighting inverse to these estimated probabilities, one estimator of
overall opinions can be formed.  An alternative estimator can be
formed by directly regressing to predict the survey outcome given any
available sources of information.  Doubly robust estimation unifies
these two techniques, so that unbiasedness is guaranteed if \emph{either} the probability
estimate is accurate \emph{or} the regressed predictor is accurate.
%then the doubly robust estimator is unbiased.

We apply the doubly robust technique to policy value
estimation in a contextual bandit setting.
The core technique is analyzed in terms of bias in
\sec{bias} and variance in~\sec{variance}. Unlike previous
theoretical analyses, we do not assume that either the reward model or
the past policy model are correct. Instead, we show how the deviations
of the two models from the truth impact bias and variance of the
doubly robust estimator. To our knowledge, this style of analysis is novel and
may provide insights into doubly robust estimation beyond the specific
setting studied here. In \sec{experiment}, we apply this method to
both policy
evaluation and optimization, 
%in both cases 
finding that this approach substantially
sharpens existing techniques. \vspace{-2mm}

\subsection{Prior Work} \vspace{-2mm}

Doubly robust estimation is widely used in statistical inference
(see, e.g., \namecite{KangSc07} and the references therein).
%~\cite{Robins94Estimation} is a statistical
%technique addressing value estimation with nonrandomized %nonrepresentative 
%sampling.
%For example, 
More recently, it has been used in Internet advertising to
estimate the effects of new features for online
advertisers~\cite{DRAds,DRAds2}. Previous work focuses on parameter
estimation rather than policy evaluation/optimization, as
addressed here.  Furthermore, most of previous analysis of doubly robust estimation
studies asymptotic behavior or relies on various modeling assumptions
(e.g., \namecite{Robins94Estimation}, \namecite{LD04}, and \namecite{KangSc07}).
%In contrast, o
Our analysis is non-asymptotic and makes no such assumptions.

Several other papers in machine learning have used ideas related to the basic
technique discussed here, although not with the same language.  For
\emph{benign bandits}, \namecite{HK} construct algorithms which
use reward estimators in order to achieve a worst-case regret that
depends on the variance of the bandit rather than time.  Similarly,
the Offset Tree algorithm~\cite{Beygelzimer09Offset} can be thought of
as using a crude reward estimate for the ``offset''.  In both cases, the
algorithms and estimators described here are substantially more
sophisticated.

\section{Problem Definition and Approach}
\label{sec:definition}

Let $\X$ be an input space and $\A=\set{1,\dotsc,k}$ a finite action
space.  A contextual bandit problem is specified by a distribution $D$
over pairs $(x,\vr)$ where $x\in\X$ is the context and
$\vr\in[0,1]^\A$ is a vector of rewards.  The input data has been
generated using some unknown policy (possibly adaptive and randomized)
as follows:

\begin{compactitem}
\item The world draws a new example $(x,\vr)\sim D$. Only $x$ is revealed.
\item The policy chooses an action $a\sim p(a\given x,h)$, where $h$
  is the history of previous observations (that is, the concatenation of all preceding contexts, actions and observed rewards).
\item Reward $r_a$ is revealed.  It should be emphasized that other rewards $r_{a'}$ with $a'\ne a$ are not observed.
\end{compactitem}

Note that neither the distribution $D$ nor the policy $p$ is known.
Given a data set $S=\set{(x,h,a,r_a)}$ collected as above,
%(history $h$ is just the concatenation of all preceding contexts, actions and rewards), 
%we consider introducing a new policy $\pi:\X\to\A$.
we are interested in two tasks: policy evaluation and policy optimization.
In policy evaluation, we are interested in estimating the \emph{value} of a stationary
policy $\pi$, defined as:
\[
  V^{\pi} = \E_{(x,\vr)\sim D}[r_{\pi(x)}]
\enspace.
\]
On the other hand, the goal of policy optimization is to find an optimal policy with maximum value:
$\pi^*=\argmax_\pi V^{\pi}$.
In the theoretical sections of the paper, we treat the problem of policy evaluation.
It is expected that better evaluation generally leads to better optimization~\cite{ESII}.
In the experimental section, we study how our policy evaluation approach can
be used for policy optimization in a classification setting.

\subsection{Existing Approaches}

The key challenge in estimating policy value, given the data as described in
the previous section, is the fact that we only have partial information about
the reward, hence we cannot directly simulate our proposed policy on
the
data set $S$. There are two common solutions for overcoming this limitation.
The first, called \emph{direct method} (DM), forms an estimate
$\hrho_a(x)$ of the expected reward conditioned on the context
\emph{and} action. The policy value is then estimated by
\[
  \hV^\pi_\DM = \frac{1}{\card{S}}\sum_{x\in S} \hrho_{\pi(x)}(x)
\enspace.
\]
\vspace{-1pt}
Clearly, if $\hrho_a(x)$ is a good approximation of the true expected
reward, defined as
  $\rho_a(x) = \E_{(x,\vr)\sim D}[r_a\given x]$,
then the DM estimate is close to $V^\pi$. Also, if $\hrho$ is unbiased, $\smash{\hat{V}}\vphantom{V}^\pi_\DM$ is an unbiased
estimate of $V^\pi$. A problem with this method is that the estimate
$\hrho$ is formed without the knowledge of $\pi$ and hence might focus
on approximating
$\rho$ mainly in the areas that are irrelevant for $V^\pi$ and not sufficiently in the areas that are
important for $V^\pi$; see \namecite{Beygelzimer09Offset} for a more refined analysis.

The second approach, called \emph{inverse propensity score} (IPS), is
typically less prone to problems with bias. Instead of
approximating the reward, IPS forms an approximation $\hp(a\given x,h)$
of $p(a\given x,h)$, and uses this estimate to correct for the shift in
action proportions between the old, data-collection policy and the new policy:
\[
  \hV^\pi_\IPS = \frac{1}{\card{S}}\sum_{(x,h,a,r_a)\in S} \frac{r_a\ind(\pi(x)=a)}{\hp(a\given x,h)}
\]
where $\ind(\cdot)$ is an indicator function evaluating to one if its
argument is true and zero otherwise. If $\hp(a\given x,h)\approx
p(a\given x,h)$ then the IPS estimate above will be, approximately, an
unbiased estimate of $V^\pi$. Since we typically have a good (or even
accurate) understanding of the data-collection policy, it is often
easier to obtain a good estimate $\hp$, and thus IPS estimator is in
practice less susceptible to problems with bias compared with the
direct method. However, IPS typically has a much larger variance, due
to the range of the random variable increasing.  The issue becomes
more severe when $p(a\given x,h)$ gets smaller.  Our approach
alleviates the large variance problem of IPS by taking advantage of
the estimate $\hrho$ used by the direct method.

\subsection{Doubly Robust Estimator}

Doubly robust estimators take advantage of both the estimate of the expected reward $\hrho_a(x)$
and the estimate of action probabilities $\hp(a\given x,h)$.  Here, we
use a DR estimator of the form first suggested by
\namecite{CasselSaWr76} for regression, but previously not studied for policy
learning:
\begin{multline}
\label{eq:DR}
 \!\!\!\!
 \!\!
 \hV^\pi_\DR = \frac{1}{\card{S}}\sum_{(x,h,a,r_a)\in S}
              \Biggl[
                     \frac{(r_a-\hrho_a(x))\ind(\pi(x)=a)}{\hp(a\given
                       x,h)}
              \Biggr.
\\[-0.5\baselineskip]
             + \hrho_{\pi(x)}(x)\Bigr].
 \!\!\!\!
\end{multline}
Informally, the estimator uses $\hrho$ as a baseline and if there is data available,
a correction is applied. We will see that our estimator is accurate if \emph{at least one}
of the estimators, $\hrho$ and $\hp$, is accurate, hence the name
\emph{doubly robust}. 

In practice, it is rare to have an accurate estimation of either
$\rho$ or $p$.
Thus, a basic question is: How does this estimator perform as the estimates
$\hrho$ and $\hp$ deviate from the truth?
%We first analyze the bias of the estimator. In the next section, we analyze its variance.
The following two sections are dedicated to bias and variance analysis,
respectively, of the DR estimator.

\section{Bias Analysis}
\label{sec:bias}

Let $\Delta$ denote the
additive deviation of $\hrho$ from $\rho$, and $\delta$ a
multiplicative deviation of $\hp$ from $p$:
\begin{align*}
  &\Delta(a,x)=\hrho_a(x)-\rho_a(x),
\\
  &\delta(a,x,h) = 1 - p(a\given x,h) / \hp(a\given x,h)
\enspace.
\end{align*}
We express the expected value of $\hV^\pi_\DR$ using $\delta(\cdot,\cdot,\cdot)$
and $\Delta(\cdot,\cdot)$. To remove clutter, we introduce shorthands
$\rho_a$ for $\rho_a(x)$, $\hrho_a$ for $\hrho_a(x)$, 
$\ind$ for $\ind(\pi(x)=a)$, $p$ for $p(\pi(x)\given x,h)$,
$\hp$ for $\hp(\pi(x)\given x,h)$, $\Delta$
for $\Delta(\pi(x),x))$, and $\delta$ for $\delta(\pi(x),x,h)$. In our
analysis,
we assume that the estimates $\hp$ and $\hrho$ are
fixed independently of $S$ (e.g., by splitting the original data set
into $S$ and a separate portion for estimating $\hp$ and
$\hrho$). To evaluate $\E[\hV^\pi_\DR]$, it suffices to focus on a
single term in \eq{DR}, conditioning on $h$:
\begin{equation}
\label{eq:decomp}
\begin{aligned}
&
 \E_{(x,\vr)\sim D, a\sim p(\cdot\given x,h)}
  \BiggBracks{
      \frac{(r_a-\hrho_a)\ind}{\hp} + \hrho_{\pi(x)}
  }
\\
  &{}=
  \E_{x,\vr,a\given h}\BiggBracks{
      \frac{(r_a-\rho_a-\Delta)\ind}{\hp}
      + \rho_{\pi(x)} + \Delta
  }
\\
 &{}=
  \E_{x,a\given h}\BiggBracks{
      \frac{(\rho_a-\rho_a)\ind}{\hp}
  +
      \Delta\bigParens{
             1-\ind/\hp}
  }
 +\E_{x}[\rho_{\pi(x)}]
 \!
\\
 &{}=
 \E_{x\given h}\bigBracks{
        \Delta\bigParens{
             1-p/\hp}}
+V^\pi
  =
\E_{x\given h}[\Delta\delta]
 +
 V^\pi
\enspace.
\end{aligned}
\end{equation}
Even though $x$ is independent of $h$, the conditioning on $h$ remains in the last line,
because $\delta$, $p$ and $\hp$ are functions of $h$. Summing across all terms in \eq{DR}, we
obtain the following theorem:

\begin{theorem}
\label{thm:DR:bias}
Let $\Delta$ and $\delta$ be defined as above. Then, the bias of the doubly robust
estimator is
\[
\bigAbs{\E_S[\hV^\pi_\DR] - V^\pi} =
\frac{1}{\card{S}}\BigAbs{\E_S\BigBracks{\sum_{(x,h)\in
      S}\Delta\delta}}
\enspace. 
\]
If the past policy and the past policy estimate are stationary (i.e.,
independent of $h$), the expression simplifies to
\begin{equation}
%\label{eq:DR:bias}
\notag
\bigAbs{\E[\hV^\pi_\DR] - V^\pi} = \bigAbs{\E_x[\Delta\delta]}
\enspace.
\end{equation}
\end{theorem}

In contrast (for simplicity we assume stationarity):
\begin{align*}
& \bigAbs{\E[\hV^\pi_\DM] - V^\pi} =\bigAbs{\E_x[\Delta]}
\\
& \bigAbs{\E[\hV^\pi_\IPS] - V^\pi} = \bigAbs{\E_x[\rho_{\pi(x)}\delta]}
\enspace,
\end{align*}
where the second equality is based on the observation that IPS is a
special case of DR for $\hrho_a(x)\equiv 0$.

In general, neither of the estimators dominates the others. However,
if \emph{either} $\Delta\approx 0$, \emph{or} $\delta\approx 0$, the
expected
value of the doubly robust
estimator will be close to the true value, whereas DM requires 
$\Delta\approx 0$ and IPS requires $\delta\approx 0$. Also, if $\Delta\approx 0$
and $\delta \ll 1$, DR will still
outperform DM, and similarly for IPS with roles of $\Delta$ and
$\delta$
reversed. Thus, DR can effectively take advantage of
both sources of information for better estimation.

\section{Variance Analysis}
\label{sec:variance}

In the previous section, we argued that the expected value of
$\hV^\pi_\DR$ compares favorably with IPS and DM. In this section, we
look at the variance of DR.  Since large deviation bounds have a
primary dependence on variance, a lower variance implies a faster
convergence rate. We treat only the case with stationary past policy, 
and hence drop the dependence on $h$ throughout.

As in the previous
section, it suffices to analyze the second moment (and then variance) of a single term of
\eq{DR}. We use a similar
decomposition as in Eq.~\eqref{eq:decomp}. To simplify derivation we use
the notation
$
  \eps=(r_a-\rho_a)\ind/\hp
$.
Note that, conditioned on $x$ and $a$, the expectation
of $\eps$ is zero. Hence, we can write the second moment as
\begin{align*}
&
  \E_{x,\vr,a}
 \BiggBracks{
  \BiggParens{\frac{(r_a-\hrho_a)\ind}{\hp} + \hrho_{\pi(x)}}^2
  }
\\
&\quad{}=
  \E_{x,\vr,a}[\eps^2]
  +\E_{x}[\rho_{\pi(x)}^2]
+2\E_{x,a}\bigBracks{
        \rho_{\pi(x)}
        \Delta\bigParens{
             1-\ind/\hp}
   }
\\
&\qquad{}
 +\E_{x,a}\bigBracks{
        \Delta^2\bigParens{
             1-\ind/\hp}^2
   }
\\
&\quad{}=
  \E_{x,\vr,a}[\eps^2]
  +\E_{x}[\rho_{\pi(x)}^2]
+2\E_{x}\bigBracks{
        \rho_{\pi(x)}
        \Delta\delta}
\\
&\qquad{}
 +\E_{x}\bigBracks{
        \Delta^2\bigParens{
             1- 2p/\hp + p/\hp^2}
  }
\\
&\quad{}=
  \E_{x,\vr,a}[\eps^2]
  +\E_{x}[\rho_{\pi(x)}^2]
  +2\E_{x}\bigBracks{
         \rho_{\pi(x)}
         \Delta\delta}
\\
&\qquad{}
 +\E_{x}\bigBracks{
        \Delta^2\bigParens{
             1- 2p/\hp + p^2/\hp^2 + p(1-p)/\hp^2}
  }
\\
&\quad{}=
  \E_{x,\vr,a}[\eps^2]
  +\E_{x}\bigBracks{\bigParens{
        \rho_{\pi(x)} + \Delta\delta}^2
    }
\\
&\qquad{}
 +\E_{x}\bigBracks{
        \Delta^2\cdot p(1-p)/\hp^2
   }
\\
&\quad{}=
  \E_{x,\vr,a}[\eps^2]
  +\E_{x}\bigBracks{\bigParens{
        \rho_{\pi(x)} + \Delta\delta}^2
    }
\\
&\qquad{}
 +\E_{x}\BiggBracks{
        \frac{1-p}{p}\cdot\Delta^2(1-\delta)^2
   }
\enspace.
\end{align*}
Summing across all terms in \eq{DR} and combining with \thm{DR:bias}, we obtain
the variance:

\begin{theorem}
\label{thm:DR:var}
Let $\Delta$, $\delta$ and $\eps$ be defined as above. If the past policy and
the policy estimate are stationary, then the variance of the doubly
robust estimator is
\begin{multline*}
\Var\bigBracks{
   \hV^\pi_\DR}
=
 \frac{1}{\card{S}}\Biggl(
 \E_{x,\vr,a}[\eps^2]
  +\Var_{x}\bigBracks{
        \rho_{\pi(x)} + \Delta\delta}
  \Biggr.
\\
\Biggl.{}
 +\E_{x}\BiggBracks{
        \frac{1-p}{p}\cdot\Delta^2(1-\delta)^2
   }
 \Biggr)
\enspace.
\end{multline*}
\end{theorem}
Thus, the variance can be decomposed into three terms. The first
accounts for randomness in rewards. The second term is the variance
of the estimator due to the randomness in $x$. And the last term can
be viewed as the importance weighting penalty. A similar expression can be
derived for the IPS estimator:
\begin{multline*}
\Var\bigBracks{
   \hV^\pi_\IPS}
=
 \frac{1}{\card{S}}\Biggl(
 \E_{x,\vr,a}[\eps^2]
  +\Var_{x}\bigBracks{
        \rho_{\pi(x)} - \rho_{\pi(x)}\delta}
  \Biggr.
\\
\Biggl.{}
 +\E_{x}\BiggBracks{
        \frac{1-p}{p}\cdot\rho_{\pi(x)}^2(1-\delta)^2
   }
 \Biggr)
\enspace.
\end{multline*}
The first term is identical, the second term will
be of similar magnitude as the corresponding term of the DR estimator,
provided that $\delta\approx 0$. However, the third term
can be much larger for IPS if $p(\pi(x)\given x)\ll1$
and $\abs{\Delta}$ is
smaller than $\rho_{\pi(x)}$. In contrast, for the direct method, we obtain
\[
\Var\bigBracks{
   \hV^\pi_\DM}
=
\frac{1}{\card{S}}
\Var_{x}\bigBracks{
        \rho_{\pi(x)} + \Delta}
\enspace.
\]
Thus, the variance of the direct method does not have terms depending
either on the past policy or the randomness in the rewards. This fact
usually suffices to ensure that it is significantly lower than the
variance of DR or IPS. However, as we mention in the previous
section, the bias of the direct method is typically much larger,
leading to larger errors in estimating policy value.

\section{Experiments}
\label{sec:experiment}

This section provides empirical evidence for the effectiveness
of the DR estimator compared to IPS and DM.  We consider two classes of problems: multiclass classification with bandit feedback in public benchmark datasets
%(Section~\ref{sec:class}) 
and estimation of average user visits to an Internet portal. %(Section~\ref{sec:visit}).

\subsection{Multiclass Classification with Bandit Feedback}
\label{sec:class}

%We first reformulate multiclass classification tasks as contextual bandit problems for empirical study.  This transformation allows us to compare IPS and DR using \emph{public} datasets for evaluating and minimizing classification error.

We begin with a description of how to turn a $k$-class classification task into a $k$-armed contextual bandit problem.  This transformation allows us to compare IPS and DR using \emph{public} datasets for both policy evaluation and learning.

\subsubsection{Data Setup}
\label{sec:class-data}

\begin{table*}
\begin{center}
\begin{tabular}{|l||c|c|c|c|c|c|c|c|c|}
\hline
Dataset & ecoli & glass & letter & optdigits & page-blocks & pendigits & satimage & vehicle & yeast \\
\hline \hline
Classes ($k$) & 8 & 6 & 26 & 10 & 5 & 10 & 6 & 4 & 10 \\
\hline
Dataset size & 336 & 214 & 20000 & 5620 & 5473 & 10992 & 6435 & 846 & 1484 \\
\hline
\end{tabular}
\end{center}
\caption{Characteristics of benchmark datasets used in Section~\ref{sec:class}.}
\label{tab:class-data}
\end{table*}

In a classification task, we assume data are drawn IID from a fixed
distribution: $(x,c)\sim D$, where $x\in\X$ is
the feature vector and $c\in\{1,2,\ldots,k\}$ is the class label.  A
typical goal is to find a classifier $\pi:\X\mapsto\{1,2,\ldots,k\}$
minimizing the classification error:
$
e(\pi)=\E_{(x,c)\sim D}\left[\ind(\pi(x)\ne c)\right].
$

Alternatively, we may turn the data point $(x,c)$ into a
cost-sensitive classification example $(x,l_1,l_2,\ldots,l_k)$, where
$l_a = \ind(a \ne c)$ is the loss for predicting $a$.  Then, a classifier $\pi$ may be interpreted as an action-selection policy, and its classification error is exactly the policy's expected loss.\footnote{When considering classification problems, it is more natural to talk about minimizing classification errors.  This loss minimization problem is symmetric to the reward maximization problem defined in Section~\ref{sec:definition}.}

To construct a partially labeled dataset, exactly one loss component for each example is observed, following the approach of \namecite{Beygelzimer09Offset}.  Specifically, given any $(x,l_1,l_2,\ldots,l_k)$, we randomly select a label $a \sim \text{\textsc{unif}}(1,2,\ldots,k)$, and then only reveal the component $l_a$.  The final data are thus in the form of $(x,a,l_a)$, 
%where $1/k$ is the probability that $a$ is chosen among the $k$ labels.  Note that this data is in 
which is the form of data defined in Section~\ref{sec:definition}.  Furthermore, $p(a\given x)\equiv1/k$ and is assumed to be known.

Table~\ref{tab:class-data} summarizes the benchmark problems adopted from the UCI repository~\cite{Asuncion07Uci}.

%\subsubsection{Algorithms}
%\label{sec:class-alg}

%Algorithms: \cite{Beygelzimer08Multiclass} \cite{Beygelzimer09Offset} \cite{McAllester11Direct}

\begin{table*}
\begin{center}
\begin{tabular}{|l||c|c|c|c|c|c|c|c|c|}
\hline
Dataset & ecoli & glass & letter & optdigits & page-blocks & pendigits & satimage & vehicle & yeast \\
\hline \hline
bias (IPS) & $0.004$ & $0.003$ & $0$ & $0$ & $0$ & $0$ & $0$ & $0$ & $0.006$ \\
bias (DR) & $0.002$ & $0.001$ & $0.001$ & $0$ & $0$ & $0$ & $0$ & $0.001$ & $0.007$ \\
bias (DM)  & $0.129$ & $0.147$ & $0.213$ & $0.175$ & $0.063$ & $0.208$ & $0.174$ & $0.281$ & $0.193$ \\
\hline
rmse (IPS) & $0.137$ & $0.194$ & $0.049$ & $0.023$ & $0.012$ & $0.015$ & $0.021$ & $0.062$ & $0.099$ \\
rmse (DR) & $0.101$ & $0.142$ & $0.03$ & $0.023$ & $0.011$ & $0.016$ & $0.019$ & $0.058$ & $0.076$ \\
rmse (DM) & $0.129$ & $0.147$ & $0.213$ & $0.175$ & $0.063$ & $0.208$ & $0.174$ & $0.281$ & $0.193$ \\
\hline
\end{tabular}
\end{center}
\caption{Comparison of results in Figure~\ref{fig:class-eval}.} \label{tbl:class-eval}
\end{table*}

\begin{table*}
\begin{center}
\begin{tabular}{|l||c|c|c|c|c|c|c|c|c|}
\hline
Dataset & ecoli & glass & letter & optdigits & page-blocks & pendigits & satimage & vehicle & yeast \\
\hline \hline
IPS (DLM) & $0.52933$ & $0.6738$ & $0.93015$ & $0.64403$ & $0.08913$ & $0.5358$ & $0.40223$ & $0.39507$ & $0.72973$ \\
DR (DLM) & $0.28853$ & $0.50157$ & $0.60704$ & $0.09033$ & $0.0831$ & $0.12663$ & $0.17133$ & $0.31603$ & $0.5292$ \\
\hline
IPS (FT) & $0.46563$ & $0.90783$ & $0.9393$ & $0.84017$ & $0.3701$ & $0.73123$ & $0.69313$ & $0.63517$ & $0.81147$ \\
DR (FT) & $0.32583$ & $0.45807$ & $0.47197$ & $0.17793$ & $0.05283$ & $0.0956$ & $0.18647$ & $0.38753$ & $0.59053$ \\
\hline
Offset Tree & $0.34007$ & $0.52843$ & $0.5837$ & $0.3251$ & $0.04483$ & $0.15003$ & $0.20957$ & $0.37847$ & $0.5895$ \\
\hline
\end{tabular}
\end{center}
\caption{Comparison of results in Figure~\ref{fig:class-opt}.} \label{tbl:class-opt}
\end{table*}

\subsubsection{Policy Evaluation}
\label{sec:class-eval}

Here, we investigate whether the DR technique indeed gives more accurate estimates of the policy value (or classification error in our context).  For each dataset:
\begin{compactenum}
\item{We randomly split data into training and test sets of (roughly) the same size;}
\item{On the training set with fully revealed losses, we run a direct loss minimization (DLM)
    algorithm of \namecite{McAllester11Direct} to obtain a classifier
    (see Appendix~\ref{app:dlm} for details). This classifier
    constitutes the policy
    $\pi$ which we evaluate on test data;}
\item{We compute the classification error on fully observed test
    data. This error is treated as the ground truth for comparing various estimates;}
\item{Finally, we apply the transformation in Section~\ref{sec:class-data} to the test data to obtain a partially labeled set, from which DM, IPS, and DR estimates are computed.} \label{step:class-eval}
\end{compactenum}

\begin{figure}[t]
\begin{center}
\vspace{1cm}
\includegraphics[angle=270,width=\scalefigs\columnwidth]{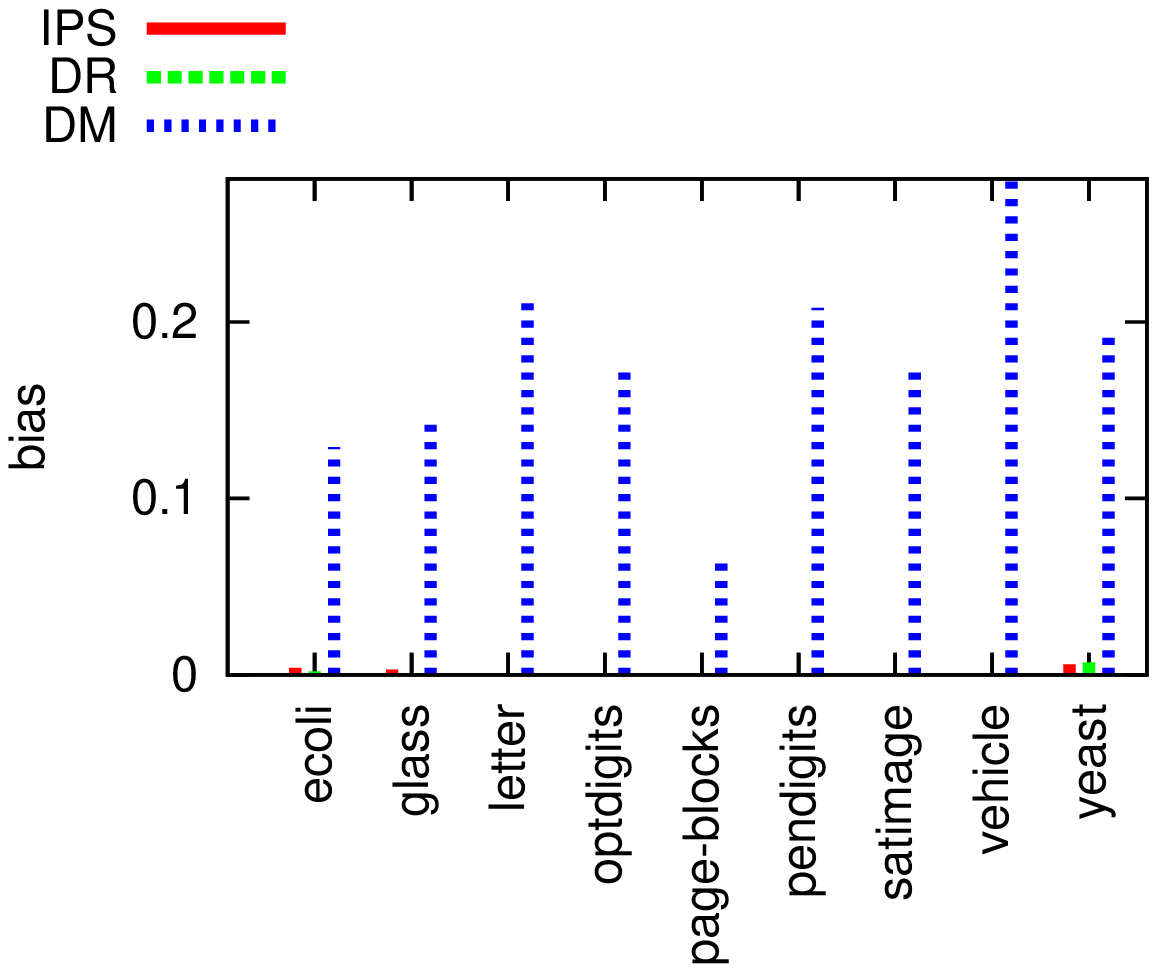} \\
\includegraphics[angle=270,width=\scalefigs\columnwidth]{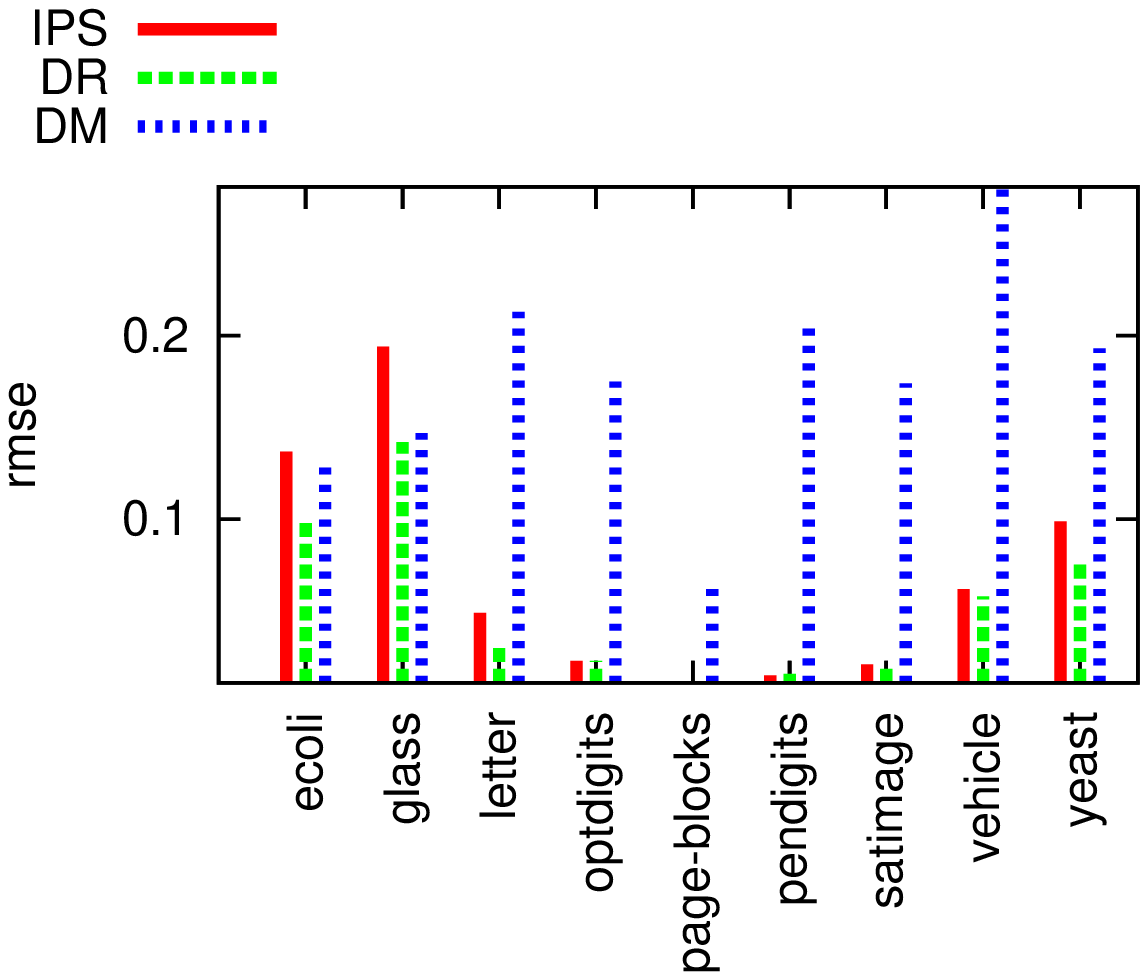}
\end{center}
\caption{Bias (upper) and rmse (lower) of the three estimators for classification error.  See Table~\ref{tbl:class-eval} for precise numbers.} \label{fig:class-eval}
\end{figure}

Both DM and DR require estimating the expected conditional loss denoted as $l(x,a)$ for given $(x,a)$.  We use a linear loss model: $\hat{l}(x,a)=w_a\cdot x$, parameterized by $k$ weight vectors $\{w_a\}_{a\in\{1,\ldots,k\}}$, and use least-squares ridge regression to fit $w_a$ based on the training set.
Step~\ref{step:class-eval} is repeated $500$ times, and the resulting
bias and rmse (root mean squared error) are reported in \fig{class-eval}.

As predicted by analysis, both IPS and DR are unbiased, since the probability estimate $1/k$ is accurate.  In contrast, the linear loss model fails to capture the classification error accurately, and as a result, DM suffers a much larger bias.

While IPS and DR estimators are unbiased, it is apparent from the rmse plot that the DR estimator enjoys a lower variance. %, which translates into a smaller rmse.  
As we shall see next, such an effect is substantial when it comes to policy optimization.

\subsubsection{Policy Optimization}
\label{sec:class-opt}

We now consider policy optimization (classifier learning).  Since DM is significantly worse on all datasets, as indicated in \fig{class-eval}, we focus on the comparison between IPS and DR.

Here, we apply the data transformation in Section~\ref{sec:class-data} to the \emph{training} data, and then learn a classifier based on the loss estimated by IPS and DR, respectively.  Specifically, for each dataset, we repeat the following steps $30$ times:
\begin{compactenum}
\item{We randomly split data into training ($70\%$) and test ($30\%$) sets;}
\item{We apply the transformation in Section~\ref{sec:class-data} to the training data to obtain a partially labeled set;}
\item{We then use the IPS and DR estimators to impute unrevealed losses in the training data;}
\item{Two cost-sensitive multiclass classification algorithms are used to learn a classifier from the losses completed by either IPS or DR: the first is DLM~\cite{McAllester11Direct}, the other is the Filter Tree reduction of \namecite{Beygelzimer08Multiclass} applied to a decision tree (see Appendix~\ref{app:tree} for more details);}
\item{Finally, we evaluate the learned classifiers on the test data to obtain classification error.} \label{step:class-opt}
\end{compactenum}

Again, we use least-squares ridge regression to build a linear loss estimator: $\hat{l}(x,a)=w_a\cdot x$.  However, since the training data is partially labeled, $w_a$ is fitted only using training data $(x,a',l_{a'})$ for which $a=a'$.

\begin{figure}[t]
\begin{center}
\vspace{1cm}
\includegraphics[angle=270,width=\scalefigs\columnwidth]{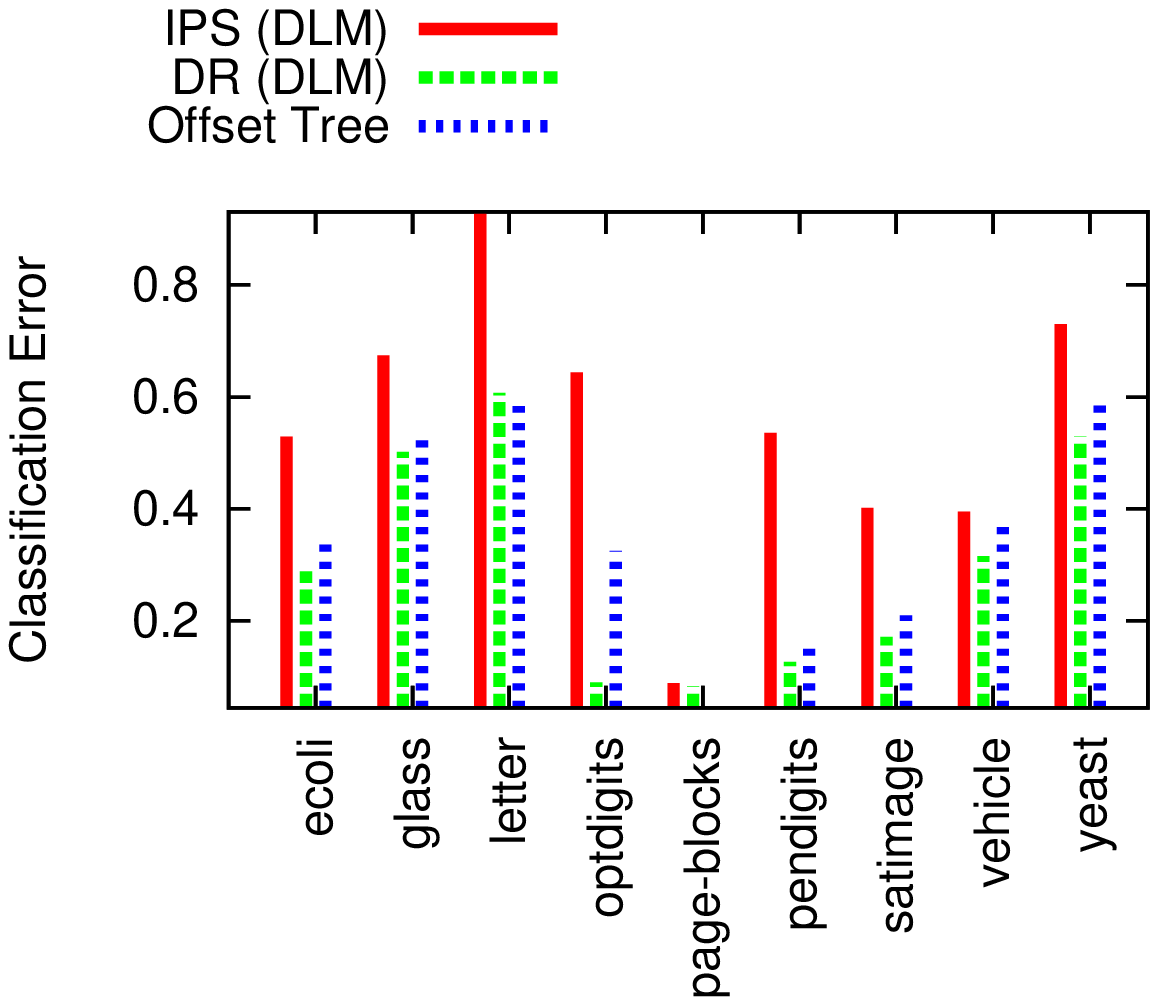} \\
\includegraphics[angle=270,width=\scalefigs\columnwidth]{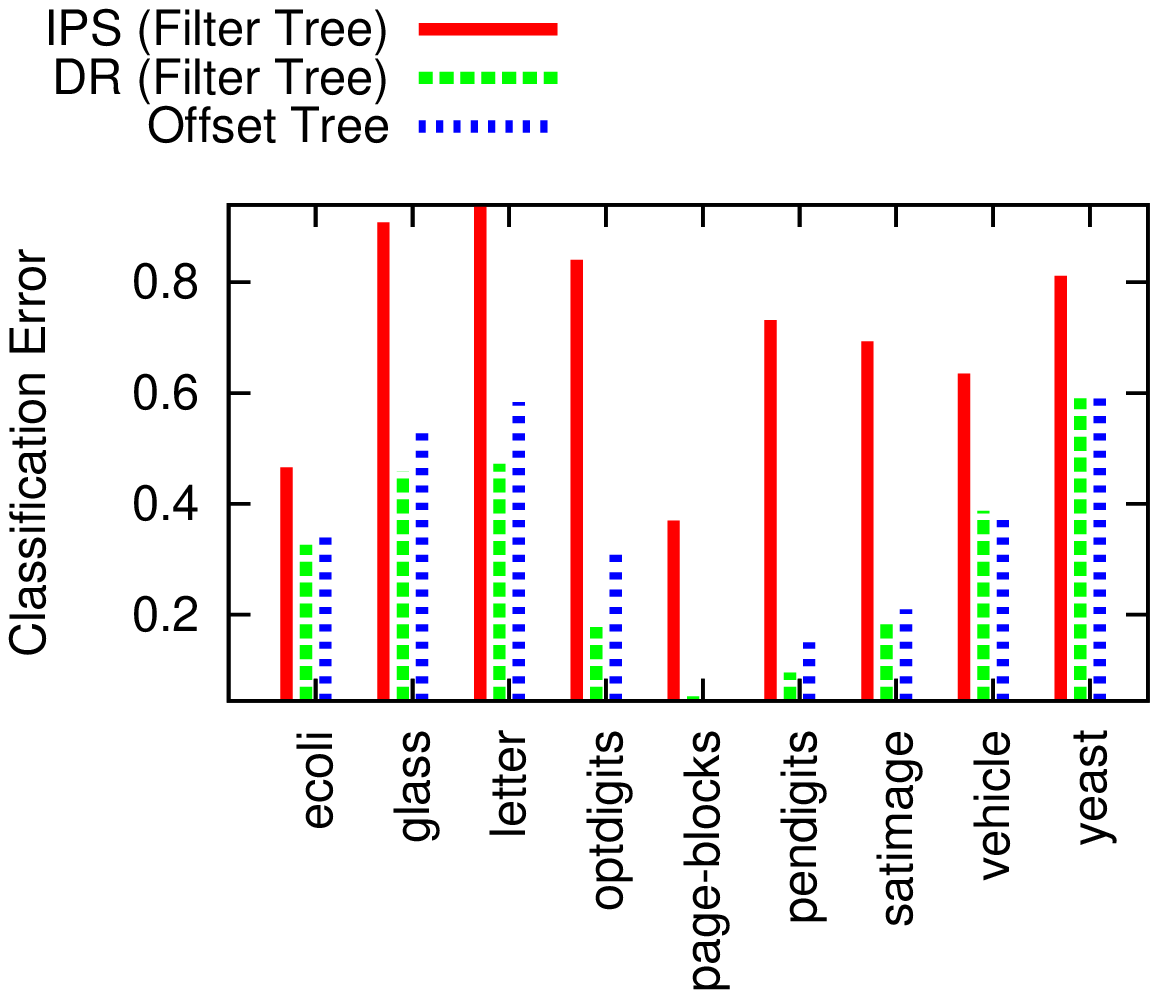}
\end{center}
\caption{Classification error of DLM (upper) and
  filter tree (lower).  Note that the representations used by DLM and
  the trees differ radically, conflating any comparison between the
  approaches.  However, the Offset and Filter Tree approaches share a
  similar representation, so differences in performance are purely a
  matter of superior optimization.
  See Table~\ref{tbl:class-opt} for precise numbers.}
\label{fig:class-opt}
\end{figure}

Average classification errors (obtained in Step~\ref{step:class-opt} above) of
the $30$ runs are plotted in \fig{class-opt}.  Clearly, for policy optimization, the advantage of the DR is even greater than for policy evaluation.  In all datasets, DR provides substantially more reliable loss estimates than IPS, and results in significantly improved classifiers.

\fig{class-opt} also includes classification error of the
Offset Tree reduction, which is designed specifically for policy
optimization with partially labeled data.\footnote{We used decision
  trees as the base learner in Offset Trees.  The numbers reported
  here are not identical to those by \namecite{Beygelzimer09Offset}
  probably because the filter-tree structures in our implementation
  were different.}  While the IPS versions of DLM and Filter Tree are
rather weak, the DR versions are competitive with Offset Tree in all
datasets, and in some cases significantly outperform Offset Tree.

Finally, we note DR provided similar improvements to two very different algorithms, one based on gradient descent, the other based on tree induction.  It suggests the generality of DR when combined with different algorithmic choices.

\subsection{Estimating Average User Visits}
\label{sec:visit}

\begin{figure}[t]
\begin{center}
\includegraphics[angle=270,width=\scalefigs\columnwidth]{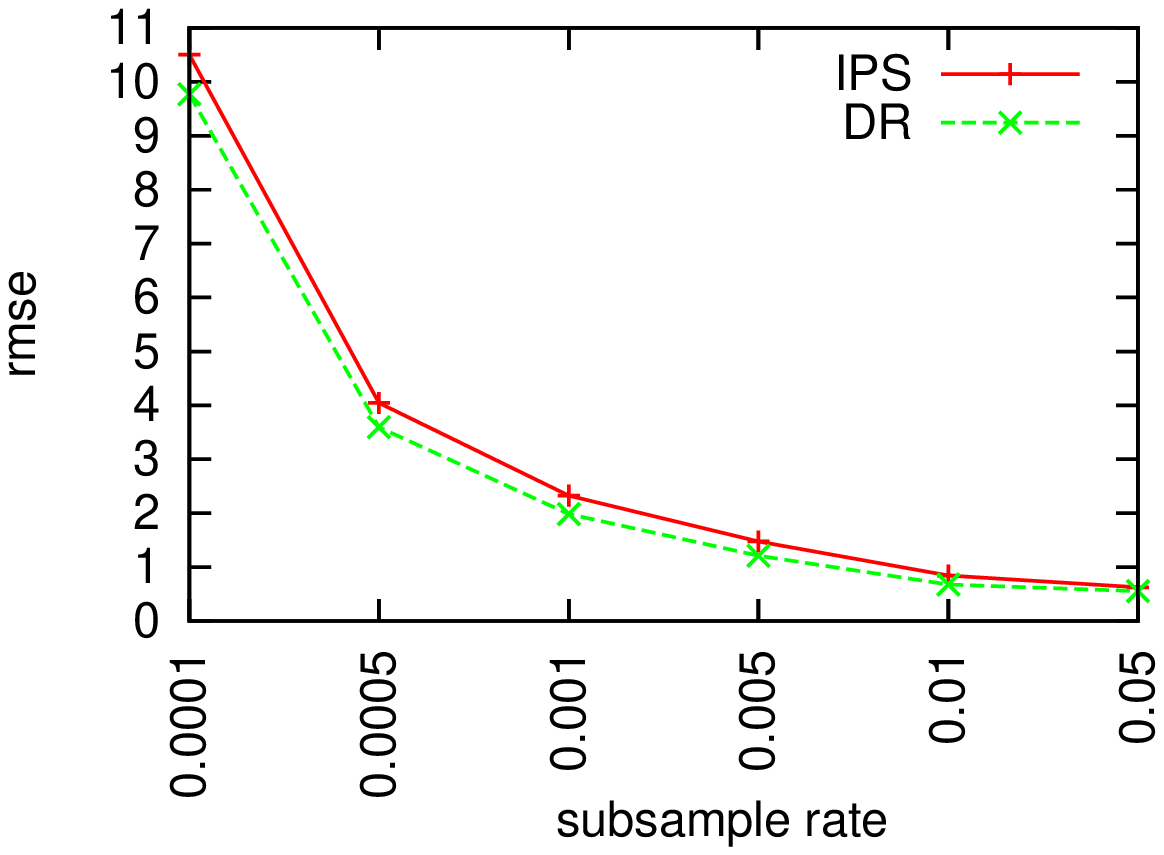} \\
\includegraphics[angle=270,width=\scalefigs\columnwidth]{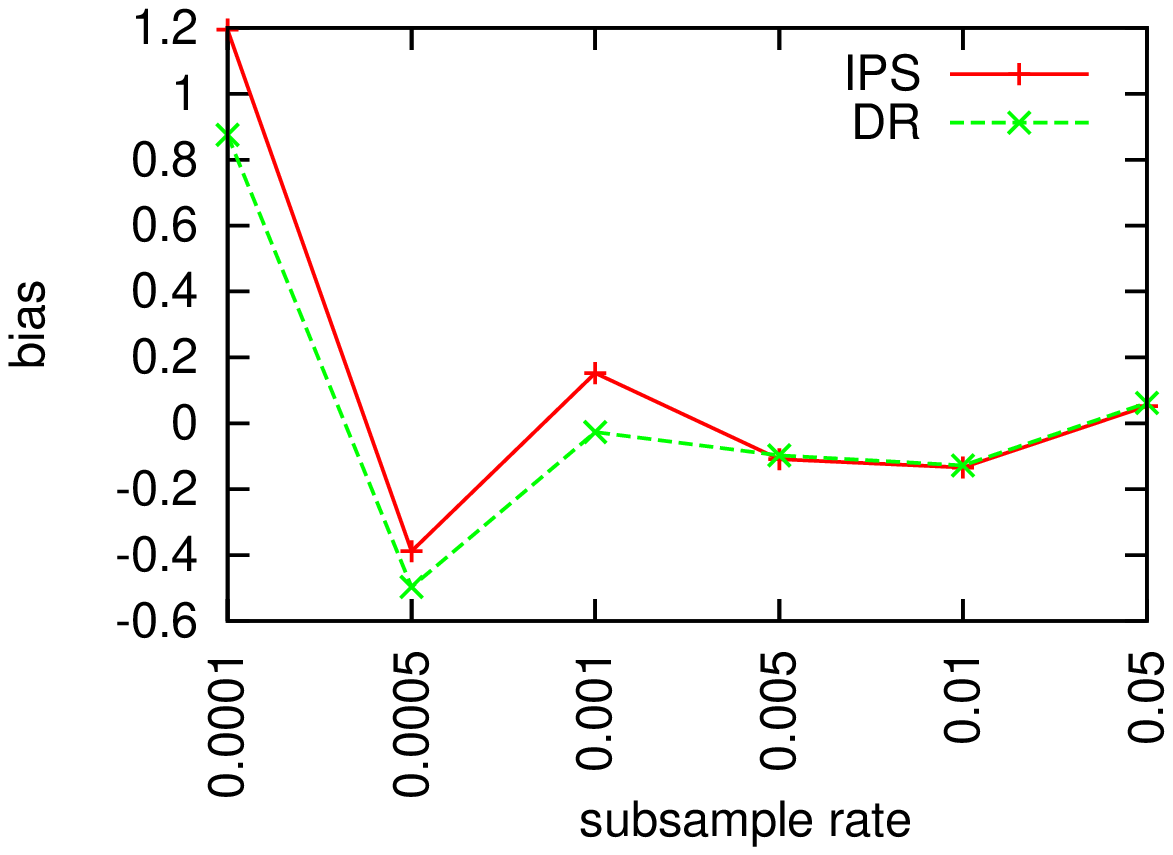}
\end{center}
\caption{Comparison of IPS and DR: rmse (top), bias (bottom).  The ground truth value is $23.8$.}
\label{fig:cv}
\end{figure}

The next problem we consider is estimating the average number of user
visits to a popular Internet portal.  Real user visits to the website
were recorded for about $4$ million \emph{bcookies}\footnote{A
bcookie is unique string that identifies a user.   Strictly speaking,
one user may correspond to multiple bcookies, but it suffices to
equate a bcookie with a user for our purposes here.} randomly
selected from all bcookies during March 2010.  Each bcookie is
associated with a sparse binary feature vector of size around $5000$.
These features describe browsing behavior as well as other information
(such as age, gender, and geographical location) of the bcookie.  We
chose a fixed time window in March 2010 and calculated the number of visits
by each selected bcookie during this window.  To summarize,
the dataset contains $N=3854689$ data: $D=\{(b_i,x_i,v_i)\}_{i=1,\ldots,N}$, where
$b_i$ is the $i$-th (unique) bcookie, $x_i$ is the corresponding
binary feature vector, and $v_i$ is the number of visits.

If we can sample from $D$ uniformly at random, the sample mean of
$v_i$ will be an unbiased estimate of the true average number of user
visits, which is $23.8$ in this problem.
However, in various situations, it may be difficult or impossible to ensure a
uniform sampling scheme due to practical constraints,
thus the sample mean may not reflect the true quantity of interest.
This is known as \emph{covariate shift}, a special case of our
problem formulated in Section~\ref{sec:definition} with $k=2$ arms.
%\textbf{Example?  ComScore?}
Formally, the partially labeled data consists of tuples $(x_i,a_i,r_i)$,
where $a_i\in\{0,1\}$ indicates whether bcookie $b_i$ is sampled,
$r_i=a_iv_i$ is the observed number of visits, and $p_i$ is the
probability that $a_i=1$.  The goal here is to evaluate the value
of a constant policy: $\pi(x)\equiv1$.

To define the sampling probabilities $p_i$, we adopted a similar approach as in \namecite{Gretton08Covariate}.  In particular, we obtained the first principal component (denoted $\bar{x}$) of all features $\{x_i\}$, and projected all data onto $\bar{x}$.  Let $\mathcal{N}$ be a univariate normal distribution with mean $m+(\bar{m}-m)/3$ and standard deviation $(\bar{m}-m)/4$, where $m$ and $\bar{m}$ were the minimum and mean of the projected values.  Then, $p_i=\min\{\mathcal{N}(x_i\cdot\bar{x}),1\}$ was the sampling probability of the $i$-th bcookie, $b_i$.

%With this large-scale, real-world data, we repeated the following steps $100$ times:
%\begin{enumerate}
%\item{We randomly subsampled a fraction $f\in\{0.0001,0.0005,0.001,0.005,0.01,0.05\}$ from the whole dataset $D$ to control data size.}
%\item{For each bcookie $b_i$ in this subsample, set $a_i=1$ with probability $p_i$, and $a_i=0$ otherwise.}
%\item{We then calculated the IPS and DR estimates on this subsample.} \label{step:visit}
%\end{enumerate}

To control data size, we randomly subsampled a fraction $f\in\{0.0001,0.0005,0.001,0.005,0.01,0.05\}$ from the entire dataset $D$.  For each bcookie $b_i$ in this subsample, set $a_i=1$ with probability $p_i$, and $a_i=0$ otherwise.  We then calculated the IPS and DR estimates on this subsample.  The whole process was repeated $100$ times.

The DR estimator required building a reward model $\hrho(x)$, which, given feature $x$, predicted the average number of visits.  Again, least-squares ridge regression was used to fit a linear model $\hrho(x)=w\cdot x$ from sampled data.

%PCA sampling trick as in the previous problem to compute the
%probability of each bcookie being subsampled.  On average, the
%subsampling probability is around $1/4$.  The set of subsampled
%bcookies is called the ``treatment group''.  Treatment groups sampled
%this way are indeed biased: in our dataset, the average number of
%visits in a treatment group is around $29.5$, in contrast to the
%actual number $23.8$, resulting in a huge bias of $24\%$.  Propensity
%scoring methods are therefore necessary to eliminate this sampling
%bias.

\fig{cv} summarizes the estimation error of the two methods with increasing data size.  For both IPS and DR, the estimation error goes down with more data.  In terms of rmse, the DR estimator is consistently better than IPS, especially when dataset size is smaller.  The DR estimator often reduces the rmse by a fraction between $10\%$ and $20\%$, and on average by $13.6\%$.  By comparing to the bias and std metrics, it is clear that DR's gain of accuracy came from a lower variance, which accelerated convergence of the estimator to the true value.  These results confirm our analysis that DR tends to reduce variance provided that a reasonable reward estimator is available.

\section{Conclusions}

Doubly robust policy estimation is an effective technique which
virtually always improves on the widely used inverse propensity score
method.  Our analysis shows that doubly robust methods tend to give
more reliable and accurate estimates.  The theory is corroborated by
experiments on both benchmark data and a large-scale, real-world
problem.

In the future, we expect the DR technique to become common practice in improving contextual bandit algorithms.  As an example, it is interesting to develop a variant of Offset Tree that can take advantage of better reward models, rather than a crude, constant reward estimate~\cite{Beygelzimer09Offset}.

\section*{Acknowledgements}

We thank Deepak Agarwal for first bringing the doubly robust technique to our attention.

\small

\appendix

%\section{Experimental Details}
%\label{app:class}

%Additional details for experiments are provided here.

\section{Direct Loss Minimization} \label{app:dlm}

Given cost-sensitive multiclass classification data $\{(x,l_1,\ldots,l_k)\}$, 
%an algorithm by \namecite{McAllester11Direct} may be instantiated 
we perform approximate gradient descent on
the policy loss (or classification error).  In the experiments of
Section~\ref{sec:class}, policy $\pi$ is specified by $k$ weight vectors $\theta_1,\ldots,\theta_k$.  Given $x\in\X$, the policy predicts as follows: $\pi(x)=\arg\max_{a\in\{1,\ldots,k\}}\{x\cdot\theta_{a}\}$.

To optimize $\theta_a$, we adapt the ``towards-better'' version of the direct loss minimization method of \namecite{McAllester11Direct} as follows: given any data $(x,l_1,\ldots,l_k)$ and the current weights $\theta_a$, the weights are adjusted by
$
\theta_{a_1} \leftarrow \theta_{a_1} + \eta x, %\qquad\qquad
\theta_{a_2} \leftarrow \theta_{a_2} - \eta x
$
where $a_1 = \arg\max_{a}\left\{x\cdot\theta_{a}-\epsilon l_{a}\right\}$, $a_2 = \arg\max_{a}\left\{x\cdot\theta_{a}\right\}$, $\eta\in(0,1)$ is a decaying learning rate, and $\epsilon>0$ is an input parameter.

For computational reasons, we actually performed batched updates rather than
incremental updatess.  We found that the
learning rate $\eta=t^{-0.3}/2$, where $t$ is the batched iteration, worked well across all datasets.  The parameter $\epsilon$ was fixed to $0.1$ for all datasets.  Updates continued until the weights converged.

Furthermore, since the policy loss is not convex in the weight vectors, we repeated the algorithm $20$ times with randomly perturbed starting weights and then returned the best run's weight according to the learned policy's loss in the training data.  We also tried using a holdout validation set for choosing the best weights out of the $20$ candidates, but did not observe benefits from doing so.

\section{Filter Tree} \label{app:tree}

The Filter Tree~\cite{Beygelzimer08Multiclass} is a reduction from
cost-sensitive classification to binary classification.  Its
input is of the same form as for Direct Loss Minimization, but its
output is a binary-tree based predictor where each node of the Filter
Tree uses a binary classifier---in this case the J48 decision tree
implemented in Weka~3.6.4~\cite{Hall09Weka}.  Thus,
there are 2-class decision trees in the nodes, with the nodes arranged
as per a Filter Tree.  Training in a Filter Tree proceeds bottom-up,
with each trained node filtering the examples observed by its parent
until the entire tree is trained.  

Testing proceeds root-to-leaf, implying that the test time
computation is logarithmic in the number of classes.  We did not test
the all-pairs Filter Tree, which has test time computation linear in
the class count similar to DLM.

\bibliography{refs}
\bibliographystyle{icml2011condensed}

\end{document}